\documentclass[a4paper,11pt]{article}
\usepackage{colacl}

\usepackage{epsfig}
\usepackage{alltt}
\usepackage{xspace}

\newcommand{\lexc}{\textbf{lexc}\xspace}
\newcommand{\twolc}{\textbf{twolc}\xspace}
\newcommand{\xerox}{Xerox\xspace}
\newcommand{\re}[1]{\texttt{#1}}
\newcommand{\fst}{{\sc fst}}

\begin{document}

\setlength\footskip{1 cm}
\pagestyle{plain}
\pagenumbering{arabic}

\title{Finite-State Non-Concatenative Morphotactics}
\author{Kenneth R. Beesley and Lauri Karttunen\\
{\tt beesley@xrce.xerox.com, karttunen@xrce.xerox.com}\\
\\
SIGPHON-2000, Proceedings of the Fifth Workshop of the ACL\\
Special Interest Group in Computational Phonology, p. 1-12,\\
August 6, 2000, Luxembourg.}

\maketitle

\thispagestyle{plain}

\abstract{
Finite-state morphology in the general tradition of the Two-Level and
\xerox implementations has proved very successful in the production of
robust morphological analyzer-generators, including many large-scale
commercial systems.  However, it has long been recognized that these
implementations have serious limitations in handling non-concatenative
phenomena.  We describe a new technique for constructing finite-state
transducers that involves reapplying the regular-expression compiler
to its own output.  Implemented in an algorithm called
compile-replace, this technique has proved useful for handling
non-concatenative phenomena; and we demonstrate it on Malay full-stem
reduplication and Arabic stem interdigitation.
}

\section{Introduction}

\begin{figure*}
\begin{verbatim}
    Lexical:  uta+ma+na-ka+p+xa+samacha-i+wa
    Surface:  uta ma n  ka p xa samach  i wa
    
              uta      = house (root)
              +ma      = 2nd person possessive
              +na      = in
              -ka      = (locative, verbalizer)
              +p       = plural
              +xa      = perfect aspect
              +samacha = "apparently"
              -i       = 3rd person
              +wa      = topic marker
\end{verbatim}
\vspace{-0.5 cm}
\caption{Aymara: \emph{utamankapxasamachiwa} = "it appears that they are in your house"}
\label{aymara}
\end{figure*}

\begin{figure*}
\begin{verbatim}
    Lexical:  Paris+mut+nngau+juma+niraq+lauq+sima+nngit+junga
    Surface:  Pari  mu  nngau juma nira  lauq sima nngit tunga

              Paris     = (root = Paris)
              +mut      = terminalis case ending
              +nngau    = go (verbalizer)
              +juma     = want
              +niraq    = declare (that)
              +lauq     = past
              +sima     = (added to -lauq- indicates "distant past")
              +nngit    = negative
              +junga    = 1st person sing. present indic (nonspecific)
\end{verbatim}
\vspace{-0.5 cm}
\caption{Inuktitut: \emph{Parimunngaujumaniralauqsimanngittunga} =
``I never said I wanted to go to Paris''}
\label{inuktitut}
\end{figure*}

Most natural languages construct words by concatenating morphemes
together in strict orders.  Such ``concatenative morphotactics'' can
be impressively productive, especially in agglutinative languages like
Aymara (Figure~\ref{aymara}\footnote{I wish to thank Stuart Newton for
this example.}) or Turkish, and in agglutinative/polysynthetic
languages like Inuktitut
(Figure~\ref{inuktitut})\cite[2]{mallon:1999}. In such languages a
single word may contain as many morphemes as an average-length English
sentence.

Finite-state morphology in the tradition of the Two-Level
\cite{koskenniemi:1983} and \xerox implementations
\cite{karttunen:1991,karttunen:1994,beesley+karttunen:book} has been
very successful in implementing large-scale, robust and efficient
morphological analyzer-generators for concatenative languages,
including the commercially important European languages and
non-Indo-European examples like Finnish, Turkish and Hungarian.
However, Koskenniemi himself understood that his initial
implementation had significant limitations in handling
non-concatenative morphotactic processes:

\vspace{-0.2 cm}
\begin{quote}
``Only restricted infixation and reduplication can be handled adequately
with the present system.  Some extensions or revisions will be
necessary for an adequate description of languages possessing
extensive infixation or reduplication'' \cite[27]{koskenniemi:1983}.
\end{quote}
\vspace{-0.2 cm}
\noindent
This limitation has of course not escaped the notice of various
reviewers, e.g. Sproat\shortcite{sproat:1992}.  We shall argue that
the morphotactic limitations of the traditional implementations are
the direct result of relying solely on the concatenation operation in
morphotactic description.

We describe a technique, within the \xerox implementation of
finite-state morphology, that corrects the limitations at the source,
going beyond concatenation to allow the full range of finite-state
operations to be used in morphotactic description.  Regular-expression
descriptions are compiled into finite-state automata or transducers
(collectively called networks) as usual, and then the compiler is
re-applied to its own output, producing a modified but still
finite-state network.  This technique, implemented in an algorithm
called {\sc compile-replace}, has already proved useful for handling
Malay full-stem reduplication and Arabic stem interdigitation,
which will be described below.  Before illustrating these
applications, we will first outline our general approach to
finite-state morphology.

\section{Finite-State Morphology}

\subsection{Analysis and Generation}

\begin{figure}[h]
\centerline{\psfig{figure=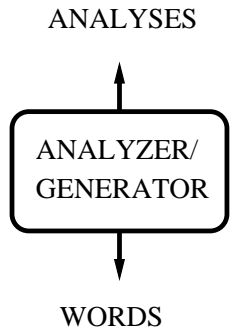}}
\vspace{-0.2 cm}
\caption{Morphological Analysis/Generation as a Relation between
Analyses and Words}
\label{relation}
\end{figure}

In the most theory- and implementation-neutral form, morphological
analysis and generation of written words can be modeled as a relation
between the words themselves and analyses of those words.
Computationally, as shown in Figure~\ref{relation}, a black-box module
maps from words to analyses to effect Analysis, and from analyses to
words to effect Generation.

\begin{figure*}[t]
\centerline{\psfig{figure=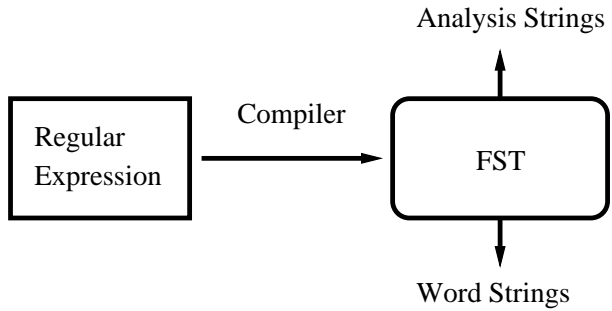}}
\caption{Compilation of a Regular Expression into an \fst~that Maps
between Two Regular Languages}
\label{fsrelation}
\end{figure*}

The basic claim or hope of the finite-state approach to
natural-language morphology is that relations like that represented in
Figure~\ref{relation} are in fact regular relations, i.e. relations
between two regular languages.  The surface language consists of
strings (= words = sequences of symbols) written according to some
defined orthography.  In a commercial application for a natural
language, the surface language to be modeled is usually a given,
e.g. the set of valid French words as written according to standard
French orthography.  The lexical language again consists of strings,
but strings designed according to the needs and taste of the linguist,
representing analyses of the surface words.  It is sometimes
convenient to design these lexical strings to show all the constituent
morphemes in their morphophonemic form, separated and identified as in
Figures~\ref{aymara} and \ref{inuktitut}.  In other applications, it
may be useful to design the lexical strings to contain the traditional
dictionary citation form, together with linguist-selected ``tag''
symbols like \re{+Noun}, \re{+Verb}, \re{+SG}, \re{+PL}, that convey
category, person, number, tense, mood, case, etc.  Thus the lexical string
representing \emph{paie}, the first-person singular, present indicative
form of the French verb \emph{payer} (``to pay''), might be spelled
\re{payer+IndP+SG+P1+Verb}. The tag symbols are stored and manipulated
just like alphabetic symbols, but they have multicharacter print
names.

If the relation is finite-state, then it can be defined using the
metalanguage of regular expressions; and, with a suitable compiler,
the regular expression source code can be compiled into a finite-state
transducer (\fst), as shown in Figure~\ref{fsrelation}, that
implements the relation computationally.  Following convention, we
will often refer to the upper projection of the \fst, representing
analyses, as the {\sc lexical} language, a set of lexical strings;
and we will refer to the lower projection as the {\sc surface}
language, consisting of surface strings. There are compelling
advantages to computing with such finite-state machines, including
mathematical elegance, flexibility, and for most natural-language
applications, high efficiency and data-compaction.

One computes with \fst s by applying them, in either direction, to an
input string.  When one such \fst~that was written for French is
applied in an upward direction to the surface word \emph{maisons}
(``houses''), it returns the related string
\re{maison+Fem+PL+Noun}, consisting of the citation form 
and tag symbols chosen by a linguist to
convey that the surface form is a feminine noun in the plural form.  A
single surface string can be related to multiple lexical strings,
e.g. applying this \fst~in an upward direction to surface string
\emph{suis} produces the four related lexical strings shown in
Figure~\ref{suis}.  Such ambiguity of surface strings is very common.

\begin{figure}[h]
\re{~~~~~~\^etre+IndP+SG+P1+Verb}\\
\verb!      suivre+IndP+SG+P2+Verb!\\
\verb!      suivre+IndP+SG+P1+Verb!\\
\verb!      suivre+Imp+SG+P2+Verb!
\vspace{-0.2 cm}
\caption{Multiple Analyses for \emph{suis}}
\label{suis}
\end{figure}

Conversely, the very same \fst~can be applied in a downward direction
to a lexical string like \re{\^etre+IndP+SG+P1+Verb} to return the related
surface string \emph{suis}; such transducers are inherently bidirectional.
Ambiguity in the downward direction is also possible, as in the relation of 
the lexical string \re{payer+IndP+SG+P1+Verb} (``I pay'') to the surface
strings \emph{paie} and \emph{paye}, which are in fact valid alternate
spellings in standard French orthography.

\subsection{Morphotactics and Alternations}

There are two challenges in modeling natural language morphology:

\begin{itemize}
\item Morphotactics
\item Phonological/Orthographical Alternations
\end{itemize}
\noindent

\begin{figure*}[ht]
\centerline{\psfig{figure=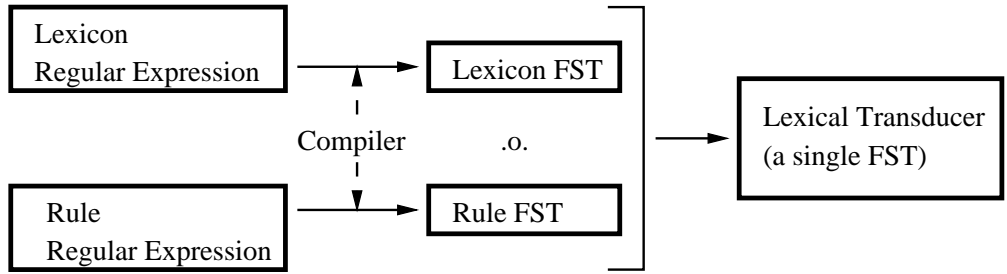}}
\caption{Creation of a Lexical Transducer}
\label{ltransducer}
\end{figure*}

Finite-state morphology models both using regular expressions.  The
source descriptions may also be written in higher-level notations
(e.g. \lexc \cite{karttunen:1993b},
\twolc\cite{karttunen+beesley:1992} and Replace Rules
\cite{karttunen:1995,karttunen:1996,kempe+karttunen:1996}) that are
simply helpful shorthands for regular expressions and that compile,
using their dedicated compilers, into finite-state networks. In
practice, the most commonly separated modules are a lexicon \fst,
containing lexical strings, and a separately written set of rule \fst s
that map from the strings in the lexicon to properly spelled surface
strings. The lexicon description defines the morphotactics of the
language, and the rules define the alternations.  The separately
compiled lexicon and rule \fst s can subsequently be composed together
as in Figure~\ref{ltransducer} to form a single ``lexical transducer''
\cite{karttunen+kaplan+zaenen:1992} that could have been defined
equivalently, but perhaps less perspicuously and less efficiently,
with a single regular expression.

\begin{figure}[h]
\re{Lexical side:}
\vspace{0.2 cm}\\
\centerline{\psfig{figure=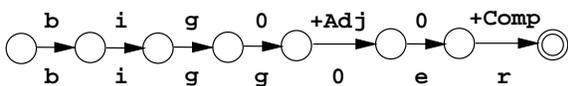}}\\
\re{Surface side:}\\
\vspace{-0.6 cm}
\caption{A Path in a Transducer for English}
\label{bigger}
\end{figure}

In the lexical transducers built at Xerox, the strings on the
lower side of the transducer are inflected surface forms of the
language. The strings on upper side of the transducer contain the
citation forms of each morpheme and any number of tag symbols that
indicate the inflections and derivations of the corresponding surface
form. For example, the information that the comparative of the
adjective \emph{big} is \emph{bigger} might be represented in the
English lexical transducer by the path (= sequence of states and arcs)
in Figure~\ref{bigger} where the zeros represent epsilon
symbols.\footnote{The epsilon symbols and their placement in the
string are not significant.  We will ignore them whenever it is
convenient.}
\noindent
The gemination of \re{g} and the epenthetical \re{e} in the surface
form \emph{bigger} result from the composition of the original lexicon
\fst~with the rule \fst~representing the regular morphological
alternations in English.

For the sake of clarity, Figure~\ref{bigger} represents the upper (=
lexical) and the lower (= surface) side of the arc label separately on
the opposite sides of the arc. In the remaining diagrams, we use a
more compact notation: the upper and the lower symbol are combined
into a single label of the form \re{upper:lower} if the symbols are
distinct. A single symbol is used for an identity pair. In the
standard notation, the path in Figure~\ref{bigger} is labeled
as\\ \centerline{\re{b i g 0:g +Adj:0 0:e +Comp:r}.}

Lexical transducers are more efficient for analysis and generation
than the classical two-level systems \cite{koskenniemi:1983} because
the morphotactics and the morphological alternations have been
precompiled and need not be consulted at runtime. But it would be
possible in principle, and perhaps advantageous for some purposes, to
view the regular expressions defining the morphology of a language as
an uncompiled ``virtual network''.  All the finite-state operations
(concatenation, union, intersection, composition, etc.) can be
simulated by an apply routine at runtime.

Most languages build words by simply stringing morphemes (prefixes,
roots and suffixes) together in strict orders.  The morpho\-tact\-ic
(word-building) processes of prefixation and suffixation can be
straightforwardly modeled in finite state terms as concatenation.
But some natural languages also exhibit non-concatenative
morphotactics.  Sometimes the languages themselves are called
``non-concatenative languages'', but most employ significant
concatenation as well, so the term ``not completely concatenative''
\cite{lavie+itai+ornan+rimon:1988} is usually more appropriate.

In Arabic, for example, prefixes and suffixes attach to stems in the
usual concatenative way, but stems themselves are formed by a process
known informally as interdigitation; while in Malay, noun plurals
are formed by a process known as full-stem reduplication.
Although Arabic and Malay also include prefixation and suffixation
that are modeled straightforwardly by concatenation, a complete
lexicon cannot be obtained without non-concatenative processes.

We will proceed with descriptions of how Malay reduplication and
Semitic stem interdigitation are handled in finite-state
morphology using the new compile-replace algorithm.

\section{Compile-Replace}

The central idea in our approach to the modeling of non-concatenative
processes is to define networks using regular expressions, as before;
but we now define the strings of an intermediate network so that they
contain appropriate substrings that are themselves in the format of
regular expressions. The compile-replace algorithm then reapplies
the regular-expression compiler to its own output, compiling the
regular-expression substrings in the intermediate network and replacing
them with the result of the compilation.

\begin{figure}[ht]
\centerline{\psfig{figure=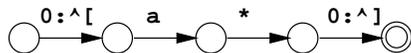}}
\vspace{-0.2 cm}
\caption{A Network with a Regular-Expression Substring
 on the Lower Side}
\label{a-star}
\end{figure}

To take a simple non-linguistic example, Figure~\ref{a-star}
represents a network that maps the regular expression \re{a*}
into \verb!^[a*^]!; that is, the same expression enclosed
between two special delimiters, \verb!^[! and \verb!^]!, that
mark it as a regular-expression substring.

\begin{figure}[ht]
\centerline{\psfig{figure=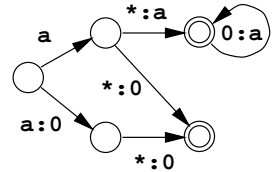}}
\vspace{-0.2 cm}
\caption{After the Application of Compile-Replace}
\label{a-star2}
\end{figure}

The application of the compile-replace algorithm to the lower side
of the network eliminates the markers, compiles the regular expression
\re{a*} and maps the upper side of the path to the language resulting
from the compilation. The network created by the operation is shown in
Figure~\ref{a-star2}.

When applied in the ``upward'' direction, the transducer in
Figure~\ref{a-star2} maps any string of the infinite \re{a*}
language into the regular expression from which the language was
compiled.

\begin{figure}[ht]
\centerline{\psfig{figure=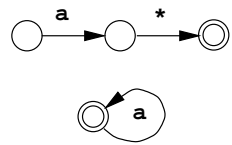}}
\vspace{-0.2 cm}
\caption{Networks Illustrating Steps 2 and 3 of the Compile-Replace
Algorithm}
\label{a-star3}
\end{figure}

The compile-replace algorithm is essentially a variant of a simple
recursive-descent copying routine. It expects to find delimited
regular-expression substrings on a given side (upper or lower) of the
network. Until an opening delimiter \verb!^[!  is encountered, the
algorithm constructs a copy of the path it is following. If the
network contains no regular-expression substrings, the result will be
a copy of the original network. When a \verb!^[!  is encountered, the
algorithm looks for a closing \verb!^]! and extracts the path between
the delimiters to be handled in a special way:
\begin{enumerate}
\item The symbols along the indicated side of the path are concatenated
  into a string and eliminated from the path leaving just the symbols
  on the opposite side. 
\item A separate network is created that contains the modified path.
\item The extracted string is compiled into a second network with the
      standard regular-expression compiler.
\item The two networks are combined into a single one using the
crossproduct operation.
\item The result is spliced between the states representing the
      origin and the destination of the regular-expression path.
\end{enumerate}
After the special treatment of the regular-expression path is
finished, normal processing is resumed in the destination state
of the closing \verb!^]! arc. For example, the result shown in
Figure~\ref{a-star2} represents the crossproduct of the two networks
shown in Figure~\ref{a-star3}.

\begin{figure*}[ht]
\begin{verbatim}                 Lexical:      b a g i   +Noun +Plural
                 Surface: ^[ { b a g i }         ^ 2   ^]

                 Lexical:      p e l a b u h a n   +Noun +Plural
                 Surface: ^[ { p e l a b u h a n }         ^ 2   ^]\end{verbatim}
\vspace{-0.5 cm}
\caption{Two Paths in the Initial Malay Transducer Defined via
Concatenation}
\label{beforereduplication}
\end{figure*}

In this simple example, the upper language of the original network in
Figure~\ref{a-star} is identical to the regular expression that is
compiled and replaced. In the linguistic applications presented in the
next sections, the two sides of a regular-expression path contain
different strings. The upper side contains morphological information;
the regular-expression operators appear only on the lower side and are
not present in the final result.

\begin{figure*}[t]
\begin{verbatim}             Lexical: b a g i         +Noun +Plural
             Surface: b a g i b a g i

             Lexical: p e l a b u h a n                   +Noun +Plural
             Surface: p e l a b u h a n p e l a b u h a n\end{verbatim}
\vspace{-0.5 cm}
\caption{The Malay \fst~After the Application of Compile-Replace to
the Lower-Side Language}
\label{afterreduplication}
\end{figure*}

\subsection{Reduplication}

Traditional Two-Level implementations are already capable of
describing some limited reduplication and infixation as in Tagalog
\cite[156--162]{antworth:1990}.  The more challenging phenomenon is
variable-length reduplication, as found in Malay and the closely
related Indonesian language.

An example of variable-length full-stem reduplication occurs with the
Malay stem \emph{bagi}, which means ``bag'' or ``suitcase''; this form
is in fact number-neutral and can translate as the plural.  Its overt
plural is phonologically \emph{bagi\-bagi},\footnote{In the standard
orthography, such reduplicated words are written with a hyphen, e.g.
\emph{bagi-bagi}, that we will ignore for this example.} formed by
repeating the stem twice in a row.  Although this pluralization
process may appear concatenative, it does not involve concatenating a
predictable pluralizing morpheme, but rather copying the preceding
stem, whatever it may be and however long it may be.  Thus the overt
plural of \emph{pelabuhan} (``port''), itself a derived form, is
phonologically \emph{pelabuhanpelabuhan}.

Productive reduplication cannot be described by finite-state or even
context-free formalisms. It is well known that the copy language,
\{\emph{ww} $|$ \emph{w} $\epsilon$ L\}, where each word contains two
copies of the same string, is a context-sensitive language. However,
if the ``base'' language L is finite, we can construct a
finite-state network that encodes L and the reduplications of all the
strings in L. On the assumption that there are only a finite number
of words subject to reduplication (no free compounding), it is
possible to construct a lexical transducer for languages such as
Malay. We will show a simple and elegant way to do this with strictly
finite-state operations.

To understand the general solution to full-stem reduplication using
the compile-replace algorithm requires a bit of background. In the
regular expression calculus there are several operators that involve
concatenation. For example, if \re{A} is a regular expression denoting
a language or a relation, \re{A*} denotes zero or more and \re{A+}
denotes one or more concatenations of \re{A} with itself.  There are
also operators that express a fixed number of concatenations.  In the
\xerox calculus, expressions of the form \verb!A^n!, where \re{n} is
an integer, denote \emph{n} concatenations of \re{A}. \re{\{abc\}}
denotes the concatenation of symbols \re{a}, \re{b}, and \re{c}.  We
also employ \verb!^[! and \verb!^]! as delimiter symbols around
regular-expression substrings.

The reduplication of any string \emph{w} can then be notated as
\verb!{w}^2!, and we start by defining a network where the lower-side
strings are built by simple concatenation of a prefix \verb!^[!, a
root enclosed in braces, and an overt-plural suffix \verb!^2!
followed by the closing \verb!^]!. Figure~\ref{beforereduplication}
shows the paths for two Malay plurals in the initial network.

The compile-replace algorithm, applied to the lower-side of this
network, recognizes each individual delimited regular-expression
substring like \verb!^[{bagi}^2^]!, compiles it, and replaces it with
the result of the compilation, here \emph{bagi\-bagi}. The same
process applies to the entire lower-side language, resulting in a
network that relates pairs of strings such as the ones in
Figure~\ref{afterreduplication}. This provides the desired solution,
still finite-state, for analyzing and generating full-stem
reduplication in Malay.\footnote{It is well-known
\cite{mccarthy+prince:1995} that reduplication can be a more complex
phenomenon than it is in Malay. In some languages only a part of the
stem is reduplicated and there may be systematic differences between
the reduplicate and the base form. We believe that our approach to
reduplication can account for these complex phenomena as well but we
cannot discuss the issue here due to lack of space.}

The special delimiters \verb!^[! and \verb!^]! can be used to surround
any appropriate regular-expression substring, using any necessary
regular-expression operators, and compile-replace may be applied to
the lower-side and/or upper-side of the network as desired. There is
nothing to stop the linguist from inserting delimiters multiple times,
including via composition, and reapplying compile-replace multiple
times (see the Appendix).  The technique implemented in
compile-replace is a general way of allowing the regular-expression
compiler to reapply to and modify its own output.

\subsection{Semitic Stem Interdigitation}

\subsubsection{Review of Earlier Work}

Much of the work in non-concatenative finite-state morphotactics has
been dedicated to handling Semitic stem interdigitation.  An example
of interdigitation occurs with the Arabic stem \re{katab}, which
means ``wrote''.  According to an influential autosegmental analysis
\cite{mccarthy:1981}, this stem consists of an all-consonant root
\re{ktb} whose general meaning has to do with writing, an abstract
consonant-vowel template \re{CVCVC}, and a voweling or vocalization
that he symbolized simply as \re{a}, signifying perfect aspect and
active voice.  The root consonants are associated with the \re{C} slots of
the template and the vowel or vowels with the \re{V} slots, producing a
complete stem \emph{katab}.  If the root and the vocalization are
thought of as morphemes, neither morpheme occurs continuously in the
stem.  The same root \re{ktb} can combine with the template
\re{CVCVC} and a different vocalization \re{ui}, signifying
perfect aspect and passive voice, producing the stem \emph{kutib},
which means ``was written''.  Similarly, the root \re{ktb} can
combine with template \re{CVVCVC} and \re{ui} to produce
\emph{kuutib}, the root \re{drs} can combine with \re{CVCVC} and
\re{ui} to form \emph{duris}, and so forth.

Kay \shortcite{kay:1987} reformalized the autosegmental tiers of
McCarthy \shortcite{mccarthy:1981} as projections of a multi-level
transducer and wrote a small Prolog-based prototype that handled the
interdigitation of roots, CV-templates and vocalizations into abstract
Arabic stems; this general approach, with multi-tape transducers, has
been explored and extended by Kiraz in several papers
\shortcite{kiraz:1994a,kiraz:1996d,kiraz:1994aa,kiraz:2000} with
respect to Syriac and Arabic. The implementation is described in
Kiraz and Grimley-Evans \shortcite{kiraz+grimley-evans:1999}.

In work more directly related to the current solution, it was Kataja
and Koskenniemi \shortcite{kataja+koskenniemi:1988} who first
demonstrated that Semitic (Akkadian) roots and patterns\footnote{These
patterns combine what McCarthy \shortcite{mccarthy:1981} would call
templates and vocalizations.} could be formalized as regular
languages, and that the non-concatenative interdigitation of stems
could be elegantly formalized as the intersection of those regular
languages.  Thus Akkadian words were formalized as consisting of
morphemes, some of which were combined together by intersection and
others of which were combined via concatenation.  

This was the key insight: morphotactic description could employ
various finite-state operations, not just concatenation; and languages
that required only concatenation were just special cases.  By
extension, the widely noticed limitations of early finite-state
implementations in dealing with non-concatenative morphotactics could
be traced to their dependence on the concatenation operation in
morphotactic descriptions.

This insight of Kataja and Koskenniemi was applied by Beesley in a
large-scale morphological analyzer for Arabic, first using an
implementation that simulated the intersection of stems in code at
runtime
\cite{beesley:1989,beesley+buckwalter+newton:1989,beesley:1990,beesley:1991},
and ran rather slowly; and later, using \xerox
finite-state technology \cite{beesley:1996,beesley:1998a}, a new
implementation that intersected the stems at compile time and
performed well at runtime.  The 1996 algorithm that intersected roots
and patterns into stems, and substituted the original roots and
patterns on just the lower side with the intersected stem, was
admittedly rather ad hoc and computationally intensive, taking over
two hours to handle about 90,000 stems on a SUN Ultra workstation.
The compile-replace algorithm is a vast improvement in both generality
and efficiency, producing the same result in a few minutes.

Following the lines of Kataja and Koskenniemi
\shortcite{kataja+koskenniemi:1988}, we could define intermediate
networks with regular-expression substrings that indicate the
intersection of suitably encoded roots, templates, and vocalizations
(for a formal description of what such regular-expression substrings
would look like, see Beesley
\shortcite{beesley:1998b,beesley:1998e}). However, the general-purpose
intersection algorithm would be expensive in any non-trivial
application, and the interdigitation of stems represents a special
case of intersection that we achieve in practice by a much more
efficient finite-state algorithm called {\sc merge}.

\subsubsection{Merge}

The merge algorithm is a pattern-filling operation that combines two
regular languages, a template and a filler, into a single one.  The
strings of the filler language consist of ordinary symbols such as
\re{d}, \re{r}, \re{s}, \re{u}, \re{i}. The template expressions may
contain special class symbols such as \re{C} (= consonant) or \re{V}
(= vowel) that represent a predefined set of ordinary symbols.  The
objective of the merge operation is to align the template strings
with the filler strings and to instantiate the class symbols
of the template as the matching filler symbols.

Like intersection, the merge algorithm operates by following two
paths, one in the template network, the other in the filler network,
and it constructs the corresponding single path in the result
network. Every state in the result corresponds to two original states,
one in template, the other in the filler. If the original states are
both final, the resulting state is also final; otherwise it is
non-final. In other words, in order to construct a successful path,
the algorithm must reach a final state in both of the original
networks. If the new path terminates in a non-final state, it
represents a failure and will eventually be pruned out.

The operation starts in the initial state of the original
networks. At each point, the algorithm tries to find all the
successful matches between the template arcs and filler arcs.
A match is successful if the filler arc symbol is included
in the class designated by the template arc symbol. The main
difference between merge and classical intersection is in
Conditions 1 and 2 below:
\begin{enumerate}
 \item If a successful match is found, a new arc is added to
       the current result state. The arc is labeled with the
       filler arc symbol; its destination is the result state that
       corresponds to the two original destinations.
 \item If no successful match is found for a given template arc,
       the arc is copied into the current result state. Its destination
       is the result state that corresponds to the destination
       of the template arc and the current filler state.
\end{enumerate}
In effect, Condition 2 preserves any template arc that does not find a
match. In that case, the path in the template network advances to
a new state while the path in the filler network stays at the
current state.

\begin{figure}[ht]
\centerline{\psfig{figure=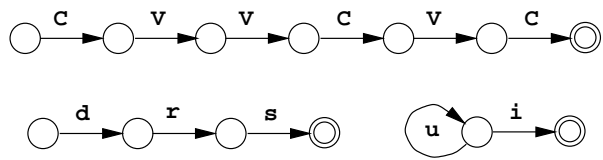}}
\vspace{-0.2 cm}
\caption{A Template Network and Two Filler Networks}
\label{merge}
\end{figure}

We use the networks in Figure~\ref{merge} to illustrate the effect
of the merge algorithm. Figure~\ref{merge} shows a linear template
network and two filler networks, one of which is cyclic.

\begin{figure}[ht]
\centerline{\psfig{figure=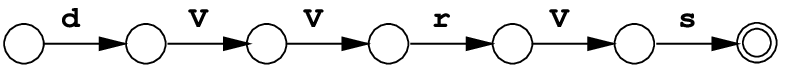}}
\vspace{-0.2 cm}
\caption{Intermediate Result.}
\label{merge2}
\end{figure}

It is easy to see that the merge of the \re{drs} network with the template
network yields the result shown in Figure~\ref{merge2}. The three
symbols of the filler string are instantiated in the three consonant
slots in the \re{CVVCVC} template.

\begin{figure}[ht]
\centerline{\psfig{figure=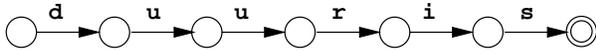}}
\vspace{-0.2 cm}
\caption{Final Result}
\label{merge3}
\end{figure}

\begin{figure*}[ht]
\begin{verbatim}
           Lexical:      k t b =Root C V C V C =Template a + =Voc
           Surface:   ^[ k t b  .m>. C V C V C    .<m.   a +      ^]

           Lexical:      k t b =Root C V C V C =Template u * i =Voc
           Surface:   ^[ k t b  .m>. C V C V C    .<m.   u * i      ^]

           Lexical:      d r s =Root C V V C V C =Template u * i =Voc
           Surface:   ^[ d r s  .m>. C V V C V C   .<m.    u * i      ^]
\end{verbatim}
\vspace{-0.5 cm}
\caption{Initial paths}
\label{beforereplacemerge}
\end{figure*}

Figure~\ref{merge3} presents the final result in which the second
filler network in Figure~\ref{merge} is merged with the intermediate
result shown in Figure~\ref{merge2}.

In this case, the filler language contains an infinite set of strings,
but only one successful path can be constructed. Because the filler
string ends with a single \re{i}, the first two V symbols can be
instantiated only as \re{u}. Note that ordinary symbols in the
partially filled template are treated like the class symbols that do
not find a match. That is, they are copied into the result in their
current position without consuming a filler symbol.

To introduce the merge operation into the \xerox regular expression
calculus we need to choose an operator symbol. Because merge, like
subtraction, is a non-commutative operation, we also must
distinguish between the template and the filler. For example, we could
choose \re{.m.} as the operator and decide by convention which of the
two operands plays which role in expressions such as \re{[A
.m.~B]}. What we actually have done, perhaps without a sufficiently
good motivation, is to introduce two variants of the merge operator,
\re{.<m.} and \re{.m>.}, that differ only with respect to whether the
template is to the left (\re{.<m.}) or to the right (\re{.m>.}) of
the filler. The expression \re{[A .<m.~B]} represents the same merge
operation as \re{[B .m>.~A]}. In both cases, \re{A} denotes the
template, \re{B} denotes the filler, and the result is the same.
With these new operators, the network in Figure~\ref{merge3} can
be compiled from an expression such as

\centerline{\re{d r s .m>.~C V V C V C .<m.~u* i}}

As we have defined them, \re{.<m.}~and \re{.m>.}~are weakly
binding left-associative operators. In this example, the
first merge instantiates the filler consonants, the second
operation fills the vowel slots. However, the order in which
the merge operations are performed is irrelevant in this case
because the two filler languages do not provide competing
instantiations for the same class symbols.

\subsubsection{Merging Roots and Vocalizations with Templates}

Following the tradition, we can represent the lexical forms of Arabic
stems as consisting of three components, a consonantal root, a \re{CV}
template and a vocalization, possibly preceded and followed by
additional affixes. In contrast to McCarthy, Kay, and Kiraz, we
combine the three components into a single projection.  In a sense,
McCarthy's three tiers are conflated into a single one with three
distinct parts. In our opinion, there is no substantive difference
from a computational point of view.

For example, the initial lexical representation of the surface forms
\emph{katab}, \emph{kutib}, and \emph{duuris}, may be represented as a
concatenation of the three components shown in
Figure~\ref{beforereplacemerge}. We use the symbols \re{=Root},
\re{=Template}, and \re{=Voc} to designate the three components of the
lexical form.  The corresponding initial surface form is a
regular-expression substring, containing two merge operators, that
will be compiled and replaced by the interdigitated surface form.

The application of the compile-replace operation to the lower side of
the initial lexicon yields a transducer that maps the Arabic
interdigitated forms directly into their corresponding tripartite
analyses and vice versa, as illustrated in
Figure~\ref{afterreplacemerge}.

Alternation rules are subsequently composed on the lower side of the
result to map the interdigitated, but still morphophonemic, strings
into real surface strings.

\begin{figure*}[t]
\begin{verbatim}
                Lexical:  k t b =Root C V C V C =Template a + =Voc
                Surface:              k a t a b

                Lexical:  k t b =Root C V C V C =Template u * i =Voc
                Surface:              k u t i b

                Lexical:  d r s =Root C V V C V C =Template u * i =Voc
                Surface:              d u u r i s
\end{verbatim}
\vspace{-0.5 cm}
\caption{After Applying Compile-Replace to the Lower Side}
\label{afterreplacemerge}
\end{figure*}

Although many Arabic templates are widely considered to be pure
CV-patterns, it has been argued that certain templates also contain
``hard-wired'' specific vowels and consonants.\footnote{ See Beesley
\shortcite{beesley:1998b} for a discussion of this controversial
issue.} For example, the so-called ``FormVIII'' template is
considered, by some linguists, to contain an embedded \re{t}:
\re{CtVCVC}.

The presence of ordinary symbols in the template does not pose any
problem for the analysis adopted here. As we already mentioned in
discussing the intermediate representation in Figure~\ref{merge2}, the
merge operation treats ordinary symbols in a partially filled template
in the same manner as it treats unmatched class symbols.  The merge of
a root such as \re{ktb} with the presumed FormVIII template and the
\re{a+} vocalism,

\centerline{\re{k t b .m>.~C t V C V C .<m.~a+}}
\noindent
produces the desired result, \re{ktatab}, without any additional
mechanism.

\section{Status of the Implementations}

\subsection{Malay Morphological Analyzer/Generator}

Malay and Indonesian are closely-related languages characterized by rich derivation
and little or nothing that could be called inflection.  The Malay
morphological analyzer prototype, written using \lexc, Replace Rules,
and compile-replace, implements approximately 50 different
derivational processes, including prefixation, suffixation,
prefix-suffix pairs (circumfixation), reduplication, some
infixation, and combinations of these processes.  Each root is marked
manually in the source dictionary to indicate the
idiosyncratic subset of derivational processes that it undergoes.

The small prototype dictionary, stored in an XML format, contains
approximately 1000 roots, with about 1500 derivational subentries
(i.e. an average of 1.5 derivational processes per root).  At compile
time, the XML dictionary is parsed and ``downtranslated'' into the
source format required for the \lexc compiler.  The XML dictionary
could be expanded by any competent Malay lexicographer.

\subsection{Arabic Morphological Analyzer/Generator}

The current Arabic system has been described in some detail in previous
publications \cite{beesley:1996,beesley:1998a,beesley:1998e} and is
available for testing on the Internet.\footnote{
http://www.xrce.xerox.com/research/mltt/arabic/}
The modification of the system to use the compile-replace algorithm
has not changed the size or the behavior of the system in any way, but
it has reduced the compilation time from hours to minutes.

\section{Conclusion}

The well-founded criticism of traditional implementations of
finite-state morphology, that they are limited to handling
concatenative morphotactics, is a direct result of their dependence on
the concatenation operation in morphotactic description.  The
technique described here, implemented in the compile-replace
algorithm, allows the regular-expression compiler to reapply to and
modify its own output, effectively freeing morphotactic description to
use any finite-state operation.  Significant experiments with Malay
and a much larger application in Arabic have shown the value of this
technique in handling two classic examples of non-concatenative
morphotactics: full-stem reduplication and Semitic stem
interdigitation. Work remains to be done in applying the technique to
other known varieties of non-concatenative morphotactics.

The compile-replace algorithm and the merge operator introduced in this
paper are general techniques not limited to handling the specific
morphotactic problems we have discussed. We expect that they will have
many other useful applications. One illustration is given in
the Appendix.

\section{Appendix: Palindrome Extraction}

To demonstrate the power of the compile-replace method, let us show
how it can be applied to solve another ``hard'' problem: identifying
and extracting all the palindromes from a lexicon. Like reduplication,
palindrome identification appears at first to require more powerful
tools than a finite-state calculus. But this task can be accomplished,
in fact quite efficiently, by using the compile-replace technique.

Let us assume that \re{L} is a simple network constructed from an
English wordlist. We start by extracting from \re{L} all the words
with a property that is necessary but not sufficient for being a
palindrome, namely, the words whose inverse is also an English
word. This step can be accomplished by redefining \re{L} as
\re{[L \& L.r]} where \re{\&} represents intersection and \re{.r}
is the reverse operator.  The resulting network contains palindromes
such as \re{madam} as well non-palindromes such as \re{dog} and
\re{god}.

The remaining task is to eliminate all the words like \re{dog} that
are not identical to their own inverse. This can be done in three
steps. We first apply the technique used for Malay reduplication.
That is, we redefine \re{L} as
 \verb!"^[" "[" L XX "]" "^" 2 "^]"!,
and apply the compile-replace operation.
At this point the lower-side of L contains strings such as \re{dogXXdogXX}
and \re{madamXXmadamXX} where \re{XX} is a specially introduced
symbol
to mark the middle (and the end) of each string.

The next, and somewhat delicate, step is to replace the \re{XX}
markers by the desired operators, intersection and reverse, and to
wrap the special regular expression delimiters
\verb!^[! and \verb!^]!  around the whole lexicon. This can be done
by composing \re{L} with one or several replace transducers to yield
a network consisting of expressions
such as \verb!^[ d o g & [d o g].r ^]! and
\verb!^[ m a d a m & [m a d a m].r ^]!

In the third and final step, the application of compile-replace
eliminates words like \re{dog} because the intersection of \re{dog}
with the inverted form \re{god} is empty. Only the palindromes survive
the operation. The extraction of all the palindromes from the 25K
Unix {\texttt /usr/dict/words} file by this method takes a couple of
seconds.

\bibliographystyle{acl}

\bibliography{big}

\end{document}